\documentclass[letterpaper, 10pt, conference]{ieeeconf}
\IEEEoverridecommandlockouts
\makeatletter
\def\endthebibliography{%
	\def\@noitemerr{\@latex@warning{Empty `thebibliography' environment}}%
	\endlist
}
\makeatother

\usepackage[utf8]{inputenc} 
\usepackage[T1]{fontenc}    
\usepackage{hyperref}       
\usepackage{url}            
\usepackage{booktabs}       
\usepackage{amsfonts}       
\usepackage{nicefrac}       
\usepackage{microtype}      
\usepackage{xcolor}         
\usepackage{graphicx} 
\usepackage{caption}
\usepackage{algorithm}
\usepackage{algpseudocode}
\usepackage{amsmath}
\usepackage{amsfonts}
\usepackage{amssymb}
\usepackage{xspace}
\usepackage{multirow}
\usepackage{bm}
\usepackage{capt-of}

\newcommand{\Tref}[1]{Table~\ref{#1}}

\newcommand{\fref}[1]{Fig.~\ref{#1}}
\newcommand{\Fref}[1]{Figure~\ref{#1}}
\newcommand{\sref}[1]{Sec.~\ref{#1}}

\newcommand{\etal}{\emph{et al.}\xspace}

\newcommand{\fcontact}{f^\text{contact}}

\title{\LARGE \bf
    Tactile Estimation of Extrinsic Contact Patch for Stable Placement
}
\author{
    Kei Ota$^{1,2}$, Devesh K. Jha$^{3}$, Krishna Murthy Jatavallabhula$^{4}$, Asako Kanezaki$^{2}$, and Joshua B. Tenenbaum$^{4}$
        \thanks{$^{1}$Kei Ota is with Mitsubishi Electric Corporation, Kanagawa, Japan. {\tt\small Ota.Kei@ds.MitsubishiElectric.co.jp}}%
	\thanks{$^{2}$Devesh K. Jha is with Mitsubishi Electric Research Labs, Cambridge, MA, USA.}%
	\thanks{$^{3}$Kei Ota and Asako Kanezaki are with Tokyo Institute of Technology, Japan.}%
	\thanks{$^{4}$Krishna Murthy Jatavallabhula and Joshua B. Tenenbaum are with Massachusetts Institute of Technology, Cambridge, MA, USA.}%
}

\begin{document}
    \twocolumn[{%
    \renewcommand\twocolumn[1][]{#1}%
    \maketitle
    \begin{center}
        \centering
        \vspace{-5mm}
        \includegraphics[width=0.95\textwidth]{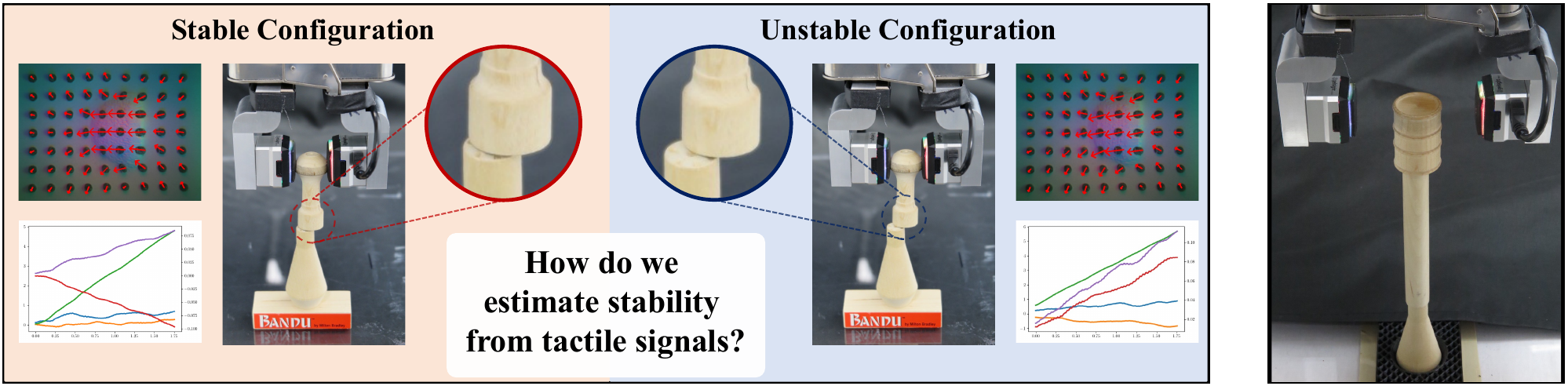}
        \captionof{figure}{\textbf{Estimating extrinsic contact from tactile sensing}: This work studies how \emph{extrinsic} contact (indirect contact between a manipulated object and an environment) can be estimated in the context of stable placement of an object in the environment with partial support.
        The figure above shows an object stacking scenario using two lightweight wooden game pieces (from the popular \emph{Bandu} puzzle).
        (Left) The contact area between the two objects being stacked is critical to the success of the stack. Vision-based tactile sensors mounted on the end effector and force-torque sensor provide us with a composite signal that includes both intrinsic (direct) and (partially observable) extrinsic contacts.
        Our key innovation is to propose a learning-based method to estimate the extrinsic contact patch using only the composite tactile signal and the knowledge of the force applied by the end effector.
        (Right) This enables the robot to stack a highly irregularly shaped object on top of a very unstable tower.}
        \label{fig:intro_fig}
    \end{center}%
    }]

    \footnotetext[1]{Kei Ota is with Mitsubishi Electric Corporation, Kanagawa, Japan. {\tt\small Ota.Kei@ds.MitsubishiElectric.co.jp}}%
    \footnotetext[2]{Devesh K. Jha is with Mitsubishi Electric Research Labs (MERL), Cambridge, MA, USA.}%
    \footnotetext[3]{Kei Ota and Asako Kanezaki are with Tokyo Institute of Technology, Japan.}%
    \footnotetext[4]{Krishna Murthy Jatavallabhula and Joshua B. Tenenbaum are with Massachusetts Institute of Technology, Cambridge, MA, USA.}%

    \thispagestyle{empty}
    \pagestyle{empty}
    \begin{abstract}
Precise perception of contact interactions is essential for fine-grained manipulation skills for robots.
In this paper, we present the design of feedback skills for robots that must learn to stack complex-shaped objects on top of each other (see \fref{fig:intro_fig}).
To design such a system, a robot should be able to reason about the stability of placement from very gentle contact interactions. Our results demonstrate that it is possible to infer the stability of object placement based on tactile readings during contact formation between the object and its environment. In particular, we estimate the contact patch between a grasped object and its environment using force and tactile observations to estimate the stability of the object during a contact formation. The contact patch could be used to estimate the stability of the object upon release of the grasp. The proposed method is demonstrated in various pairs of objects that are used in a very popular board game.
\end{abstract}
    \IEEEpeerreviewmaketitle
    \section{Introduction}\label{sec:intro}
Humans can perform very complex and precise manipulation tasks effortlessly.
Consider, for example, gently stacking two lightweight objects on top of each other without looking at them, as shown in \fref{fig:intro_fig}.
Successful execution of this task requires the object not to fall upon release of the grasp.
In these scenarios, stability is not directly observable; it must be implicitly inferred from tactile signals that entangle both \emph{intrinsic} (direct) contact between the end effector and the grasped object and \emph{extrinsic} (indirect) contact between the grasped object and the environment.
For example, in \fref{fig:intro_fig}, it is difficult to distinguish the stability of the configuration on the left from the right by looking at it visually. 
This work is motivated by how humans can disentangle a composite tactile signal to determine the nature of extrinsic contact; and can further predict whether a given stack configuration is stable.
We present a closed-loop system that similarly reasons about object stability using tactile signals that arise out of extrinsic contacts.


The stability of the object could be estimated from the contact forces experienced by an object during placement. The stability of an object is governed by the relative location of the environmental contact and the center of mass location of the object. The forces observed by the force-torque (F/T) sensor mounted on the wrist of the robot, as well as the deformation observed by the tactile sensors co-located at the gripper fingers, depend on the contact patch between the object and its environment, as well as the geometric and physical properties of the object. As a simplification, we assume that the geometry of the objects is fixed, so the robot works with known pieces. Under this assumption, the problem of estimating the stability of placement from tactile observations is simplified. With this understanding, we try to estimate the contact patch between the object and the environment using tactile signals. However, estimating contact patches from a single tactile observation is a partially observable problem. Thus, a perfect estimate of the contact from a single interaction is impossible.

To solve the partial observability problem, we present a method for aggregating information from multiple observations. The proposed method collects tactile observations by interacting with the environment multiple times and updates its belief in the underlying contact formation. We show that we can monotonically improve our estimate of the contact formation between the environment and the grasped object. This estimate is used to move the object towards a stable configuration so that it can be released in a stable pose. This is demonstrated using several pairs of objects from a popular board game where the objective is to incorporate a new block on an existing tower without destabilizing it. We also perform ablations to understand which sensing modality, the F/T sensor or the vision-based tactile sensor is helpful in understanding the phenomena during the considered contact phenomena.

\textbf{Contributions:} In summary, our contributions are the following.
\begin{enumerate}
    \item We present a method for estimating extrinsic contact patches from end-effector tactile signals that compose both intrinsic and extrinsic contacts.
    \item Our probabilistic filtering approach for use in a feedback control loop can stably stack a set of extremely challenging real-world objects using solely tactile sensing.
\end{enumerate}
    \begin{figure*}
        \centering
        \includegraphics[width=0.9\textwidth]{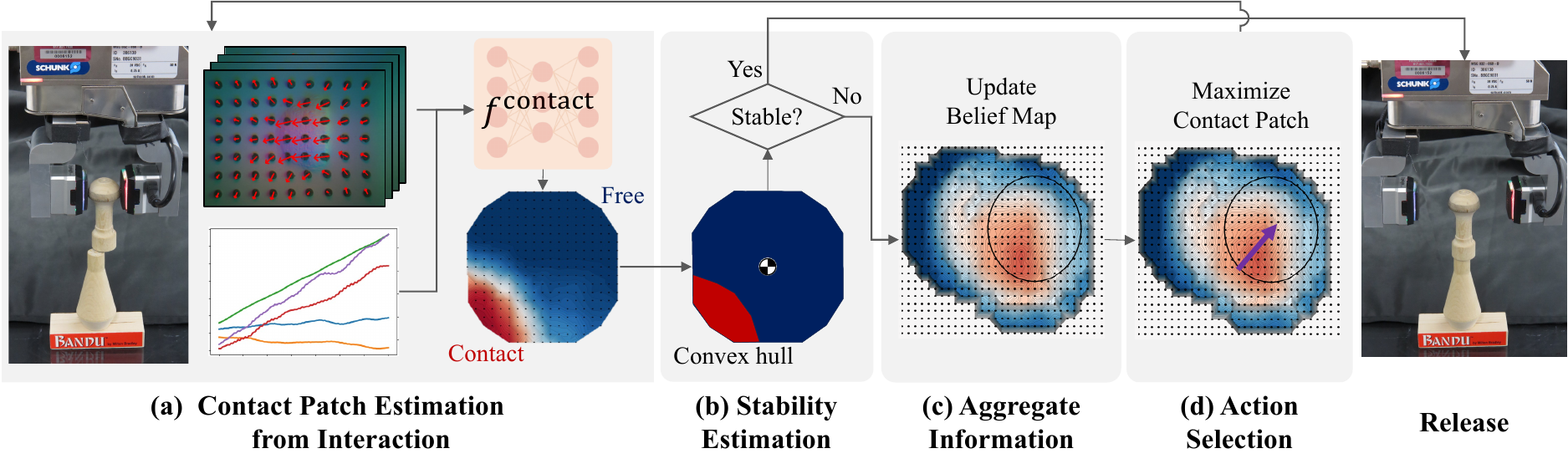}
        \caption{\textbf{Pipeline}: Our method comprises four components. First, a robot probes the environment to establish contact between the grasped object and the target object upon which it must be stacked.
        During this probing phase, we acquire a sequence of force/torque measurements and tactile images.
        We then estimate the extrinsic contact patch and, in turn, the potential stability of the resultant configuration.
        Subsequently, we aggregate the information from multiple interactions to update the belief map of the contact state.
        We pick the action that maximizes the contact patch between the objects.
        }
        \vspace{-5mm}
        \label{fig:pipeline}
    \end{figure*}
    \section{Related Work}
\textbf{Block stacking.}
Block stacking is one of the most widely studied problems in robotics. Several studies have addressed the problem of robot stacking through various approaches. These include learning to schedule auxiliary tasks for reinforcement learning (RL)~\cite{learning2018riedmiller}, combining demonstrations and RL~\cite{zhu2018reinforcement,cabi2020scaling}, employing sim-to-real transfer~\cite{zhu2018reinforcement,hermann2020adaptive,jeong2020self}, and using task-and-motion planning~\cite{noseworthy2021active}. The focus of these works primarily revolves around stacking simple cubes.
Lee \etal~\cite{lee2022beyond} propose a benchmark that introduces relatively irregular rectangles generated by deforming cubes. However, these objects still maintain convexity and simplicity. Furrer \etal~\cite{furrer2017autonomous} and Liu \etal~\cite{yifang2021planning} have explored the stacking of irregular stones. Another related work that discusses vision-based contact support could be found in~\cite{kroemer2018kernel}, however, this assumed access to the geometry of the object and was indeed reasoning about the relative placement between blocks given the object geometries. Nevertheless, these studies make assumptions regarding knowledge of geometry and assume that objects possess wide support and high friction, simplifying the problem and enabling basic pick-and-place strategies. Most importantly, these works do not reason about stability using contact information but rather perform placement using open-loop controllers. These pick-and-place stackings would not work if there is ambiguity in the location of the environment (for example, the scenario shown in Fig.~\ref{fig:intro_fig}). 
To address this problem, our proposed method considers the local contact phenomenon in which the object can topple and fall if it is not placed with the proper support. Moreover, we remove assumptions regarding the geometry of the underlying objects, necessitating the estimation of stability through interactions.

\textbf{External contact localization}
Prior works represent contacts as a set of points~\cite{manuelli2016localizing, kim2023simultaneous} and lines~\cite{ma2021extrinsic,kim2022active}. Although line contacts give us more information compared to point contacts, they require active exploration involving changes in gripper orientation~\cite{ma2021extrinsic,kim2022active}, making it difficult to apply them in our setting where the tower is very unstable.
The closest work to ours is the neural contact fields (NCF) of Higuera et al.~\cite{higuera2023neural}, where the authors estimate the contact patch between a grasped object and its environment. While NCF is evaluated on a simulation and a limited number of objects, we tested our method on unknown geometries of the environment, which can be used for an appropriate downstream task in a real system.

    \section{Problem Statement} \label{sec:problem_statement}

We are interested in performing a stable placement in environments where the object might have partial support for placement. Consider, for example, the scenario shown in \fref{fig:intro_fig}, where it is not enough to establish contact with the bottom piece but rather to estimate the object's stability in the resulting contact formation. Thus, we consider the problem of estimating the stability of an object when in contact with its environment in an attempt to release and place the object in a stable pose during a task. This is a partially observable task, as we cannot observe the full state of the system, and thus, stability needs to be estimated from sensor observations. We assume that the robot has access to tactile sensors co-located at the gripper fingers and a Force/Torque (F/T) sensor at the wrist. A certain contact formation is stable if the object can remain stable after being released from the grasp.

The stability of a contact formation depends on the relative position of the center of mass of the object and the contact patch between the object and the environment.
However, this cannot be directly observed during a contact formation, and thus leads to partial observability. 
A robot can usually observe force-torque signals and/or tactile images during interaction. The observed signals depend not only on the contact formation but also on the geometry and physical parameters of the grasped object. Thus, 
although these data have a lot of information, these are all entangled, and thus it is very difficult to extract specific information, e.g., estimate contact patch. The stability estimation problem in its full scope requires reasoning about the sensor observations while considering the geometric information of the objects. 
To simplify the estimation problem, we make the following assumptions to limit the scope of the current study.
\begin{enumerate}
    \item Geometry and physical parameters of the grasped objects are fixed.
    \item All objects are rigid and have flat surfaces.
\end{enumerate}
It is important to emphasize that the robot is unfamiliar with the shape of the underlying objects and needs to explore a stable configuration through several probing attempts. These assumptions restrict the use of our proposed objects to known objects. A full and in-depth study of the problem is left as a future exercise.
    \newcommand{\otac}{o^\text{Tac}}
\newcommand{\oft}{o^\text{FT}}

\section{Method}
This work addresses the primary challenge of estimating stability during placing irregular objects. Since the contact formation between a grasped object and the environment generates sensor observations, we estimate the contact patch between them from force and tactile measurements. 
We propose a framework consisting of four key components. 
First, the robot estimates the contact patch between the grasped object and its environment from an observation obtained by interacting with the environment. 
Then, it assesses stability based on the estimated contact patch; and releases the grasped object if it believes the current configuration is stable; otherwise, it aggregates information from multiple estimated contact patches to predict a belief map, which gives us a sense of the contact surface of the environment. Finally, the robot selects an action that moves the grasped object to a position that it believes can improve stability. In this section, we describe these four modules in more detail.

\subsection{Contact Patch Estimation} \label{sec:method_contact_patch}
Given the observed tactile image $\otac$ and F/T measurements $\oft$, our objective is to learn a model that generates a probabilistic contact patch $\hat{S}$, which consists of a set of probabilities indicating which part of the grasped object is in contact.

\begin{figure}[t]
    \centering
    \includegraphics[width=0.9\columnwidth]{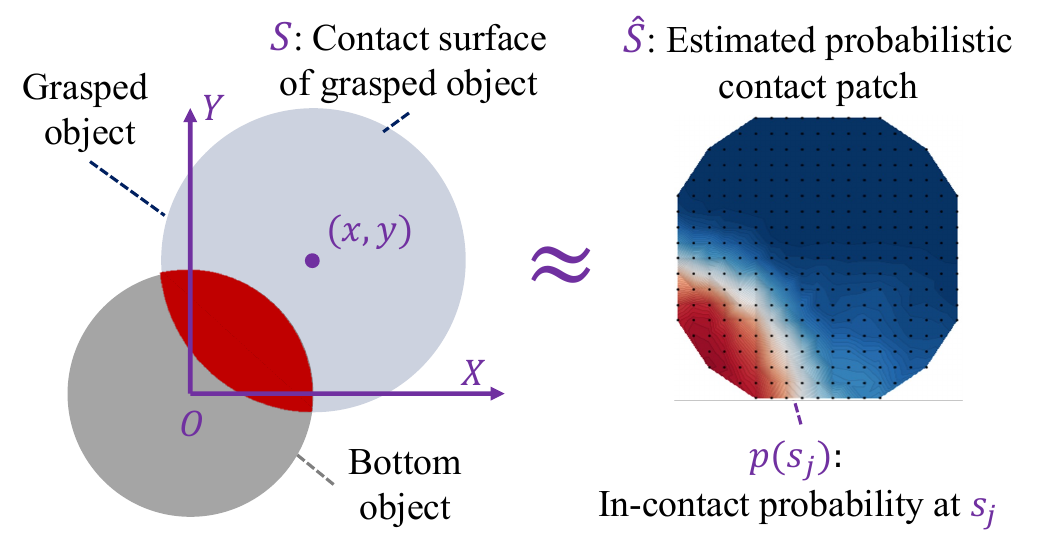}
    \caption{Definition of the \textbf{probabilistic contact patch}. (Left) The displacement $(x, y)$ is added from the origin of the bottom object $O$ during data collection. This displacement and known contact surfaces of the two objects give the ground-truth contact surface $S$. (Right) The discretized contact patch $\hat{S}$ consists of a set of probabilities $p(s_j)$ that represents whether a specific position $s_j$ of the contact surface of the grasped object is in contact or not.}
    \label{fig:definition}
    \vspace{-3mm}
\end{figure}

\textbf{Contact representation.}
To estimate the contact patch, we discretize the contact surface of the grasped object $S$ into $N$ points as $S \approx \{s_1,..., s_N \}$ each of which corresponds to a specific location on the contact surface of the grasped object (see \fref{fig:definition} right). For each point $s_j$, we predict the probability of being in contact or remaining uncontacted $p(s_j)$. Consequently, we represent the probabilistic contact patch $\hat{S}$ as a set of probabilities $\hat{S}_i = \{ p(s_1), ..., p(s_N) \}$.

\textbf{Data collection by interaction.}
During a duration of $T$ seconds, the robot applies a downward force along the negative Z axis for $d$ mm, while collecting $\otac, \oft$ from tactile and force-torque sensors at a frequency of $10$ Hz. Specifically, $\otac = \{ o^\text{Tac}_{t}\}_{t=0}^T$, where $o^\text{Tac}_{t} \in \mathbb{R}^{252}$ with $252=2\times2\times7\times9$, where we use two tactile sensors mounted on each finger and measure marker displacements on the $XY$ axis in the tactile image, and $7 \times 9$ is the number of markers in column and row (see \fref{fig:pipeline}), which can be obtained by post-processing the tactile image $I^\text{Tac}_t$. Similarly, $\oft = \{ o^\text{FT}_{t}\}_{t=0}^T$, $o^\text{FT}_{t} \in \mathbb{R}^6$ is the F/T measurement. We use a suitable impedance control to prevent the object from falling by using excessive force.
In the data collection process, we add displacements in the $XY$ plane such as $x \sim \{ x_\mathtt{min}, x_\mathtt{max} \}$ and $y \sim \{ y_\mathtt{min}, y_\mathtt{max} \}$ whose origin is the center position of the contact surface of the lower object $O$ (see \fref{fig:definition}), and the minimum and maximum ranges are defined to ensure contact between the flat surfaces of the upper and lower objects. We use known geometries and displacements to generate ground-truth contact patches for training a model.

\textbf{Training.}
Finally, we train a contact patch estimation model $\fcontact$ that takes observation $\otac, \oft$ and learns to generate a probabilistic contact surface $\hat{S}$ as:
\begin{equation}
    \hat{S} = \fcontact (\otac, \oft).
\end{equation}
This model is trained by minimizing the binary cross-entropy loss for each data point $s_j$. We use LSTM~\cite{hochreiter1997long} with two layers, each having $256$ units, to build the model to capture patterns in time-series data.

\subsection{Stability Estimation}
We utilize the estimated contact patch $\hat{S}$ to estimate the stability of the current configuration. To do that, we first construct a convex hull $C \in \text{Convex}(\hat{S})$ (see \fref{fig:pipeline} (b)) using points whose associated probability exceeds a predefined threshold denoted by $\delta$, which we use $\delta=0.9$ for our experiments. Subsequently, we check that the convex hull includes the position of the center of mass of the grasped object. In the affirmative case, the gripper releases the grasped object. Otherwise, the gripper aggregates information and moves towards a stable position by an action selection strategy described in the following sections.

\subsection{Aggregating Information from Multiple Interactions}
Since the estimation of the contact patch from tactile signals is a partially observable task, that is, multiple different contact patches can yield similar tactile signals, it is difficult to reliably estimate the contact patch from a single interaction. Therefore, we aggregate information from multiple interactions to disambiguate the estimate.

We denote the aggregated contact patch at the time step $i$ as $\hat{S}_i^B$, again representing a probabilistic contact surface of the \emph{bottom} object $\hat{S}_i^B = \{ p(s_{1,i}^B), ..., p(s_{M,i}^B)\}$, where $M$ is the number of discrete points. Following Ota \emph{et al.}~\cite{ota2023tactile}, the probabilistic formulation of the contact $p(s_i^B)$ (note we remove lowercase $m$ for simplification) given past observation and action can be formulated as
\begin{equation}
    \begin{split}
    p(s_i^B | &o_{1:i-1}, a_{1:i-1}) \\
    &= \int p(s_{i}^B | s_{i-1}^B, a_{i-1}) p(s_{i-1}^B | o_{1:i-1}, a_{1:i-1}),
    \end{split} \label{eq:prob_dynamics}
\end{equation}
where the first term is $1$ as we assume deterministic dynamics, and the second term is initialized with the prior distribution and can be obtained through recursion.
The posterior can be computed as:
\begin{equation}
    p(s_i^B | o_{1:i}, a_{1:i-1}) \propto p(o_{i} | s_i^B) p(s_i^B | o_{1:i-1}, a_{1:i-1}),
\end{equation}
where the first term is given by the contact patch estimation model $\fcontact$ and the second term can be computed from Eq.\eqref{eq:prob_dynamics}. Specifically, we initialize the probability with $p(s_0^B) = \text{Bernoulli}(0.5)$ since we do not know whether the specific point is in contact or not before interaction.

\subsection{Action Selection}
To realize a stable configuration, we design a policy that maximizes the contact surface area in the next step. The policy begins by calculating the central position of the convex hull of the aggregated contact patch $s_{C^B} = \cfrac{\sum_{i \in |C^B|} s_i^B}{|C^B|}$, where $C^B$ is again the convex hull of the aggregated contact map, and subsequently directs the robot to navigate in the direction to this central position from the current position. Furthermore, to mitigate large movement at each step, we restrict movement within $d^\text{move}$ mm if the norm exceeds $d^\text{move}$. We specifically set $d^\text{move}=3$ mm.

    \section{Experiments}
\subsection{Settings}
\textbf{Tactile sensor.}
We use a commercially available GelSight Mini tactile sensor~\cite{gelsightmini2023}, which provides 320×240 compressed RGB images at a rate of approximately 25 Hz, with a field of view of 18.6 × 14.3 millimeters. We use gels that have 63 tracking markers.

\textbf{Robot platform.}
The MELFA RV-5AS-D Assista robot, a collaborative robot with 6 DoF, is used in this study. The tactile sensor is mounted on the WSG-32 gripper (see~\fref{fig:pipeline}). We use a Force-Torque (F/T) sensor which is mounted on the wrist of the robot and used two-fold. First, we collect force observations that are used as input to the contact patch estimation model $\fcontact$. Second, the stiffness control of the position-controlled robot.

\begin{figure}[t]
	\centering
	\includegraphics[width=\columnwidth]{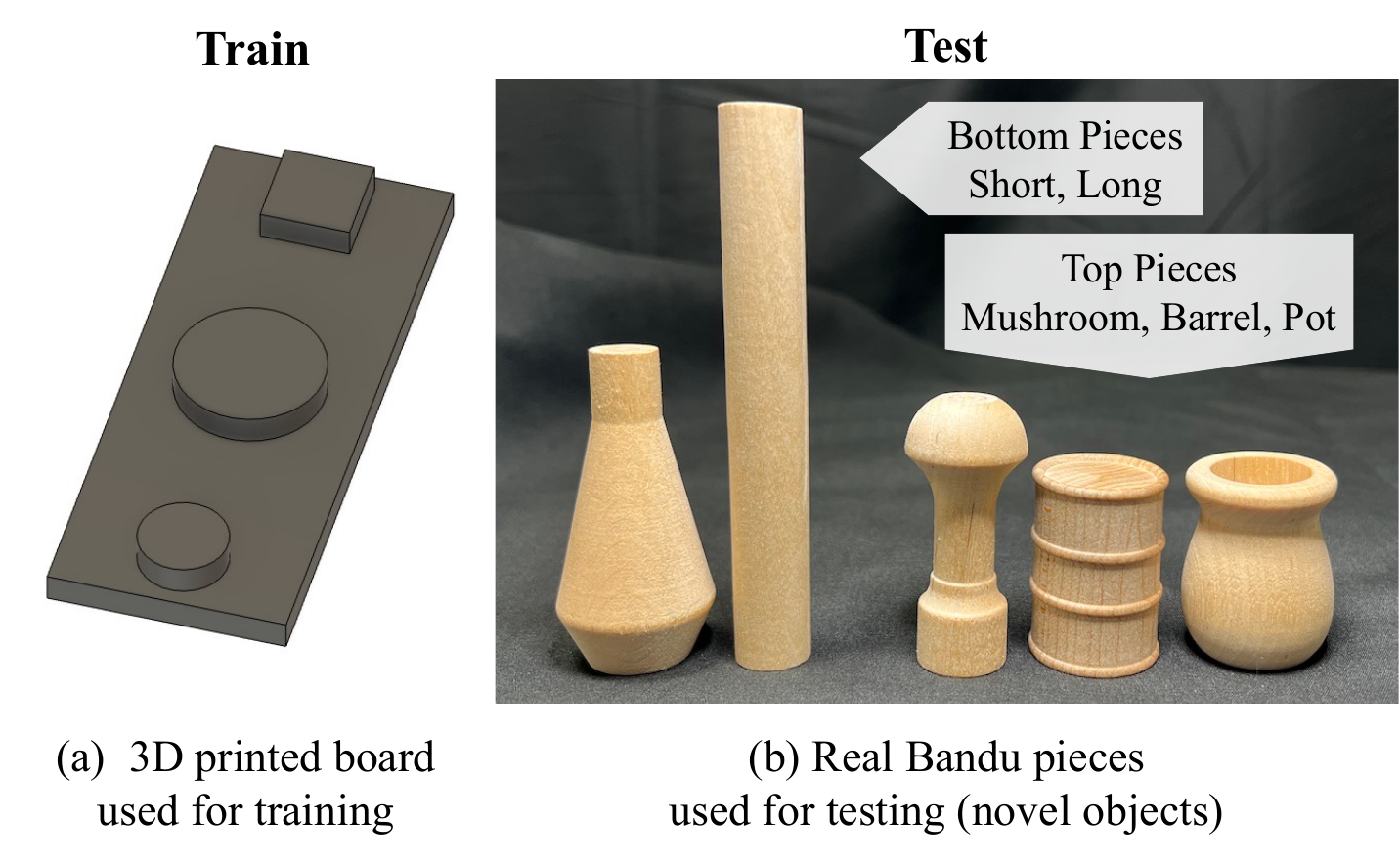}
        \caption{The 3D printed board and Bandu pieces used in our experiments. (a) We use the 3D printed board for training data collection. The board includes small and large circles with diameters of $15$ and $25$ mm and one square whose length is $15$ mm. (b) The first two pieces on the left serve as the bottom objects (or the environment), while the subsequent three on the right are designated as the grasped (top) objects. These pieces have been assigned the following names: \textit{Short}, \textit{Long}, \textit{Mushroom}, \textit{Barrel}, and \textit{Pot} from left to right.
        }
	\label{fig:bandu_piece}
\end{figure}

\textbf{Bandu.}
We use pieces from \textit{Bandu} for our experiment. Bandu is a toy game that involves stacking objects onto a base plate. Players take turns stacking these objects and compete to see who can stack the most. Each piece has a highly irregular shape, which requires robots to estimate stable placements based on the shape of the objects. \Fref{fig:bandu_piece} illustrates the Bandu pieces used in our experiments. The challenge in the game is to accommodate an irregular piece in an existing tower without destabilizing it.

\begin{figure*}[t]
    \centering
    \includegraphics[width=0.9\linewidth]{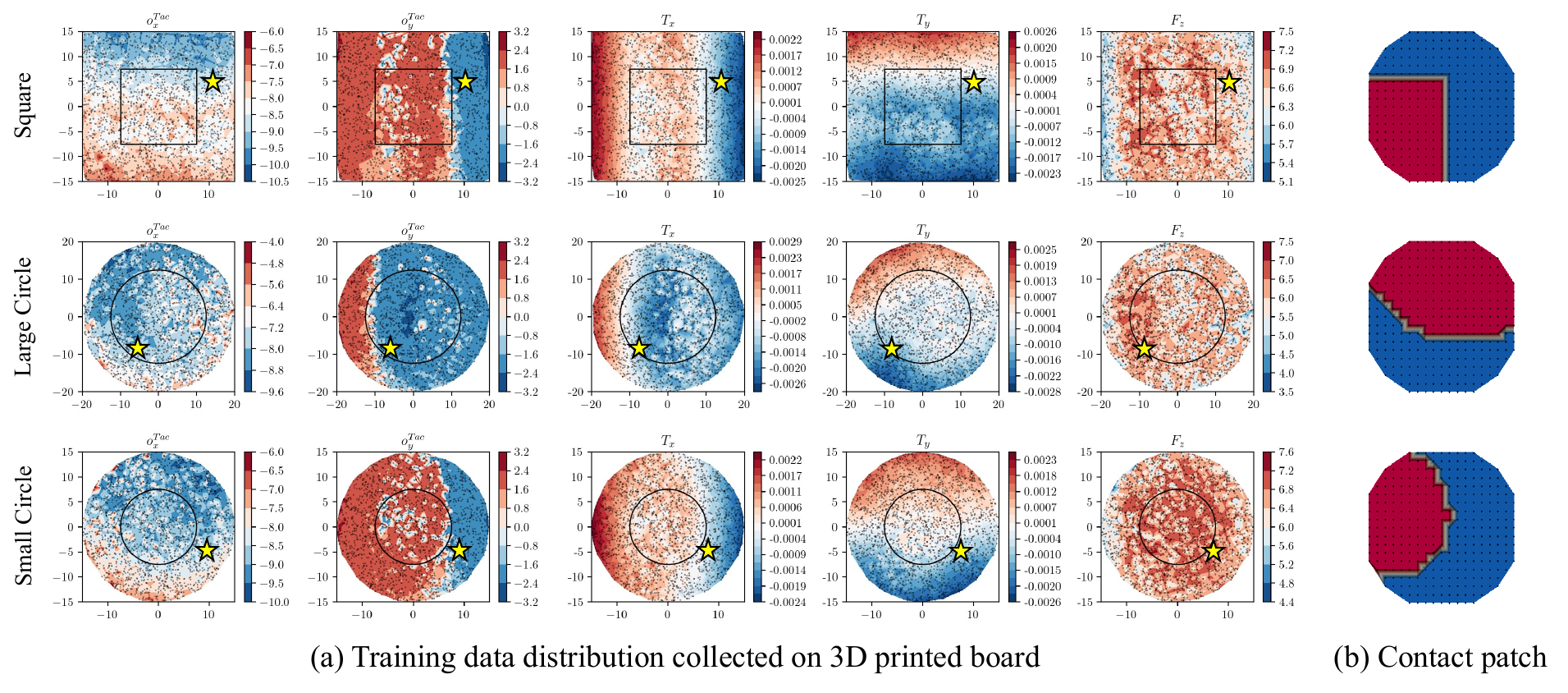}
    \vspace{-1mm}
    \caption{\textbf{Distribution of contact patches}:
    (a) Training data distribution with \textit{Pot} as the grasped object and three different 3D printed shapes as the bottom objects (see \fref{fig:bandu_piece}). Each row shows the data obtained from different primitive shapes and each column shows the distribution of different data types: tactile displacements on the $XY$ axes (only shows the maximum absolute values from all 63 tracking markers), moments on the $XY$ axes and force $F_z$. The horizontal and vertical axes show the displacements randomly added during data collection (see \fref{fig:definition}), and the black circle or rectangle in each graph shows the contour of the bottom object.
    (b) Example contact patch sampled from the star points ($\bigstar$) in the left distributions. Although these contact patches are very different, the tactile signals look quite similar as seen in the data around the star point, showing the difficulty of the task; i.e., similar tactile signals can lead to very different contact patches.
    }
    \vspace{-3mm}
    \label{fig:data_distribution}
\end{figure*}

\subsection{Data Collection} \label{subsec:exp_data_collection}
\textbf{Settings.}
We first show the distribution of the observed tactile signals to understand the difficulties of the task. We collect $2000$ tactile signals for each pair of top-bottom objects to train the contact patch estimation model $\fcontact$ by interacting with three objects on the 3D printed board as shown in \fref{fig:bandu_piece} (a), resulting in $6000$ training samples. 
During data collection, we add random displacements on the $XY$ axis as defined in \fref{fig:definition}, and let the robot go down for $d=1.5$ mm after establishing contact with the bottom object for $T=2$ seconds using the stiffness controller whose gain parameter is $(K_x, K_y, K_z)=(30,30,15)$ [N/mm]. We use the grasping force of $10$ [N].

\textbf{Results and Analysis.}
\Fref{fig:data_distribution} shows the data distribution (left) and example contact patches (right). From the first to the fourth columns, we can observe the inherent difficulties of the estimation task. In many cases, we do not observe any symmetric distribution of $\otac_x$, $\otac_y$ and the moment measurements $T_x, T_y$ about $X=0$ or $Y=0$. This could possibly be attributed to the inaccuracy in the 3D printing of the board or the slip of the object in the grasp during the contact interaction.
\fref{fig:data_distribution} (b) shows three contact patches sampled from the star positions in each row. While tactile signals near the star positions are very similar, the resulting contact patches are very different. This highlights the partial observability of the underlying contact formation, indicating that a single tactile observation may not be sufficient to localize the contact formation. This ambiguity makes training of a machine learning model very difficult because similar inputs (i.e., tactile observations) can lead to totally different outputs (i.e., contact patches).

\begin{table}
    \caption{Comparison of the contact patch estimation performance on different input modalities measured by IoU and binary classification accuracy. Bold numbers show the best results among the three different input modalities. The \textit{S} and \textit{L} of the bottom objects correspond to the \textit{Short} and \textit{Long} objects, respectively (see \fref{fig:bandu_piece}).}
    \centering
    \scalebox{0.9}{
    \begin{tabular}{rrcccccc} \toprule
        && \multicolumn{2}{c}{Mushroom} & \multicolumn{2}{c}{Barrel} & \multicolumn{2}{c}{Pot} \\
        && S & L & S & L & S & L \\ \midrule
        \multirow{3}{*}{IoU}
        & FT & $27.4$ & $37.7$ & $29.6$ & $\bm{44.5}$ & $23.8$ & $53.2$ \\
        & Tac & $33.6$ & $22.2$ & $31.5$ & $42.6$ & $16.5$ & $37.5$ \\
        & FT+Tac & $\bm{38.4}$ & $\bm{50.7}$ & $\bm{31.9}$ & $41.2$ & $\bm{24.8}$ & $\bm{54.8}$ \\ \midrule
        \multirow{3}{*}{Acc}
        & FT & $67.4$ & $65.5$ & $75.3$ & $73.1$ & $68.0$ & $71.9$ \\
        & Tac & $72.9$ & $60.4$ & $76.5$ & $\bm{77.5}$ & $66.9$ & $71.7$ \\
        & FT+Tac & $\bm{77.9}$ & $\bm{73.8}$ & $\bm{77.0}$ & $75.4$ & $\bm{67.3}$ & $\bm{74.9}$ \\
        \bottomrule
    \end{tabular}
    }
    \label{tab:comparison_input}
    \vspace{-6mm}
\end{table}

\subsection{Contact Patch Estimation}
\textbf{Settings.}
Next, we compare the performance of the contact patch estimation on different input modalities. We train the model $\fcontact$ for each top object using the dataset collected in \sref{subsec:exp_data_collection}, and we evaluate the model using the intersection-over-union (IoU) and binary classification metric. We compare the performance with three different input modalities, a F/T sensor, tactile sensors, and the combination of the two denoted as \textit{FT}, \textit{Tac}, and \textit{FT+Tac}, respectively. 
The evaluation is carried out using \emph{unseen} two Bandu pieces (see \fref{fig:bandu_piece}), which we denote as \textit{Short} and \textit{Long}. We used a 3D-printed jig to ensure that the robot always grasps the same position of the top object and collected $400$ interactions with random displacements.

\textbf{Results and Analysis.}
The results are presented in~\Tref{tab:comparison_input}. When comparing the three modalities, we can clearly see that the combination of tactile sensors and the F/T sensor (\textit{FT+Tac}) yields the best performance. Consequently, for our subsequent experiments, we will utilize both of these modalities.
However, it should be noted that the model is not confident enough to estimate the contact patch. This is because the same tactile signals can lead to different contact patches, as discussed in \sref{subsec:exp_data_collection}. Therefore, in the next experiment, we will aggregate information from multiple interactions and compare performance in stability estimation.

\subsection{Stability Estimation}
\textbf{Settings.} 
Next, we assess the stability estimation performance of the proposed method. We reuse the same data as used in the previous experiments with an additional binary label indicating whether the current configuration is stable by checking whether the geometric center of the bottom surface of the grasped object (i.e., the projection of the center of mass of the grasped object on the bottom surface) lies inside the contact patch.
We compare our method with a baseline model that directly produces the stability probability by replacing the final layer of $\fcontact$ with a fully connected layer with a single unit and sigmoid activation. We name it \textit{Implicit} because it implicitly estimates stability, while our framework explicitly predicts it through the estimated contact patch.

\begin{figure}[t]
    \centering
    \includegraphics[width=\columnwidth]{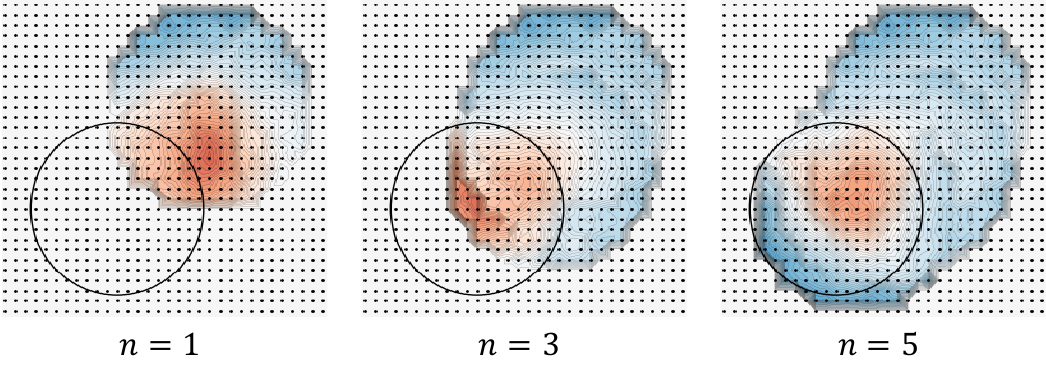}
    \caption{An example of how the proposed method aggregates multiple estimations and updates contact probability map. The circle in a solid line shows the ground-truth contour of the bottom object. While the initial estimate ($n=1$) is incorrect, the estimation accuracy monotonically improves with multiple interactions ($n=3, 5$).
    }
    \label{fig:qual_results_stab_change}
\end{figure}

\textbf{Results and Analysis.}
\Tref{tab:results_stab} shows the qualitative results. Single interaction leads to poor performance, as seen in the results of the baseline (\textit{Implicit}) as well as our method with single interaction (\textit{Ours $n=1$}).
However, by aggregating the estimates of multiple interactions, the stability estimation performance improves significantly, leading to an average accuracy of $90$\%.
\Fref{fig:qual_results_stab_change} shows how the probability of a contact patch changes during interactions. It shows that the method corrects the initial inaccurate estimate and improves accuracy with additional interactions, and the method finally reconstructs the contact surface of the bottom object with reasonable accuracy.

\begin{table}
    \caption{Stability estimation performance measured by binary accuracy. $n$ indicates the number of interactions and bold numbers show the best results.}
    \centering
    \scalebox{0.97}{
    \begin{tabular}{rrcccccc} \toprule
        && \multicolumn{2}{c}{Mushroom} & \multicolumn{2}{c}{Barrel} & \multicolumn{2}{c}{Pot} \\
        && S & L & S & L & S & L \\ \midrule
        \multicolumn{2}{c}{Implicit} & $73.7$ & $73.0$ & $63.9$ & $69.8$ & $64.2$ & $66.0$ \\ \midrule
        \multirow{3}{*}{Ours} 
        & $n=1$ & $69.8$ & $81.0$ & $63.7$ & $71.4$ & $61.7$ & $69.4$ \\
        & $n=2$ & $88.4$ & $93.0$ & $92.7$ & $88.9$ & $83.2$ & $83.9$ \\ 
        & $n=3$ & $\bm{93.0}$ & $\bm{93.2}$ & $\bm{97.3}$ & $\bm{92.1}$ & $\bm{91.1}$ & $\bm{85.7}$ \\
        \bottomrule
    \end{tabular}
    }
    \label{tab:results_stab}
\end{table}

\subsection{Stacking}
\textbf{Settings.} 
Finally, we evaluate the stacking performance of the method. We always initialize the first interaction from an unstable contact state (i.e., the object would topple upon release of grasp). We run the method $10$ times for each piece and evaluate whether the robot successfully places the piece in a stable configuration. Furthermore, we also test the method in a harder scenario, where the \textit{Long} piece is already stacked onto the \textit{Short} piece (see \fref{fig:bandu_piece} for the definition of the pieces), and we stack a top piece on top of these two objects.
We compare our method with a \textit{Pick \& Place} baseline, where it releases the piece without estimating the stability.

\begin{figure}[t]
    \centering
    \includegraphics[width=\columnwidth]{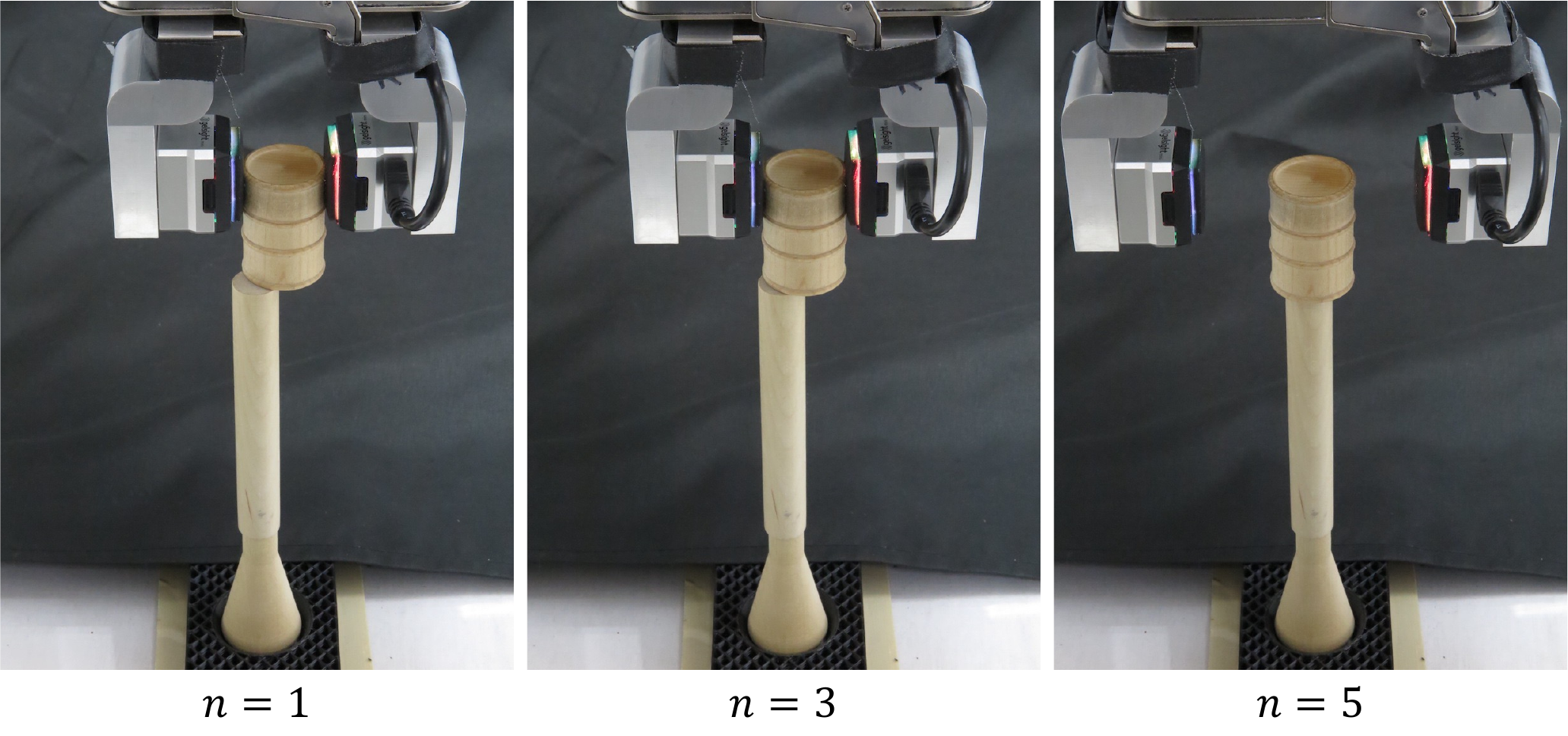}
    \caption{The robot moves towards a stable configuration and successfully stacks the \textit{Barrel} piece on top of an already built tower consisting of \textit{Short} and \textit{Long}.}
    \label{fig:qual_results_real_stacking}
\end{figure}

\textbf{Results and Analysis.}
\Tref{tab:stacking_performance} shows the results. The pick-and-place baseline fails in all trials. The proposed method improves performance by predicting the contact patch at each iteration and aggregating information to improve the estimation accuracy. Although the success rate drops when the number of bottom objects is increased, the method can still succeed with a success rate of around $60$\%. \Fref{fig:qual_results_real_stacking} shows a qualitative result of how it moves to the more stable position.

\begin{table}[t]
    \caption{Success rate of stacking. One and Two means stacking on top of a single and two objects, respectively.}
    \centering
    \scalebox{0.95}{
    \begin{tabular}{rrcccccc} \toprule
        && \multicolumn{2}{c}{Mushroom} & \multicolumn{2}{c}{Barrel} & \multicolumn{2}{c}{Pot} \\
        && S & L & S & L & S & L \\ \midrule
        \multirow{2}{*}{One}
        & Pick \& Place & $0/10$ & $0/10$ & $0/10$ & $0/10$ & $0/10$ & $0/10$ \\
        & Ours          & $8/10$ & $6/10$ & $8/10$ & $5/10$ & $7/10$ & $6/10$ \\ \midrule
        Two
        & Ours          & $6/10$ & $5/10$ & $7/10$ & $5/10$ & $6/10$ & $5/10$ \\ \bottomrule
    \end{tabular}
    }
    \label{tab:stacking_performance}
\end{table}

    \section{Conclusion}
Designing systems that can interpret and disentangle useful contact information from observed tactile measurements is the key to precise and fine manipulation.
We proposed a framework for estimating extrinsic contact patches from tactile and force-torque measurements. Contact patch estimation allows us to estimate the stability of the placement of several different objects in novel and unstable environments. We tested the proposed approach for the placement of several pieces of the game of Bandu, which is known to be a difficult stacking task. In the future, we would like to improve the performance by training on a wider variety of objects and relaxing the assumption of the known geometry so that the trained model can be used for the stacking task with arbitrary objects.

    \clearpage
    \bibliographystyle{IEEEtran}
    \bibliography{reference}
\end{document}